\documentclass[pmlr]{jmlr}
\RequirePackage{graphicx}
 \usepackage{booktabs}
 \usepackage{natbib}
 \usepackage{cleveref}
 \usepackage{hyperref}
 \usepackage[most]{tcolorbox}
\usepackage{longtable}
\usepackage{pifont}
\usepackage{ulem}
\usepackage{listings}
\makeatletter
\def\set@curr@file#1{\def\@curr@file{#1}}
\makeatother
\usepackage[load-configurations=version-1]{siunitx}

\theorembodyfont{\upshape}
\theoremheaderfont{\scshape}
\theorempostheader{:}
\theoremsep{\newline}

\jmlrproceedings{PMLR}{Proceedings of Machine Learning Research}
\jmlrvolume{340}
\jmlryear{2026}
\jmlrworkshop{Machine Learning for Healthcare}
\title[]{A Cloud--Edge System for Multimodal Clinical Screening in Resource-Constrained Rural Settings}
\author{%
  \Name{Hei Ting \nametag{(Una)} Chan\nametag{$^{1}$}} \Email{hei@umich.edu}\\
  \Name{Chenwei Wu\nametag{$^{1}$}} \Email{chenweiw@umich.edu}\\
  \Name{Xueshen Liu\nametag{$^{1}$}} \Email{liuxs@umich.edu}\\
  \Name{Zesen Zhao\nametag{$^{1}$}} \Email{hymanzzs@umich.edu}\\
  \Name{Boyuan Zheng\nametag{$^{1}$}} \Email{boyuann@umich.edu}\\
  \Name{Luis Filipe Nakayama\nametag{$^{2}$}} \Email{luisnaka@mit.edu}\\
  \Name{Michael G. Morley\nametag{$^{3}$}} \Email{michael\_morley@meei.harvard.edu}\\
  \Name{Liyue Shen\nametag{$^{1}$}} \Email{liyues@umich.edu}\\
  \Name{Jiasi Chen\nametag{$^{1}$}} \Email{jiasi@umich.edu}\\
  \Name{Z. Morley Mao\nametag{$^{1}$}} \Email{zmao@umich.edu}\\
  \addr $^{1}$University of Michigan, Ann Arbor, MI, USA\\
  \addr $^{2}$Massachusetts Institute of Technology, Cambridge, MA, USA\\
  \addr $^{3}$Harvard Medical School, Boston, MA, USA}
\begin{document}
\maketitle
\begin{abstract}
Medical AI has demonstrated specialist-level diagnostic accuracy, yet these capabilities remain largely inaccessible in resource-constrained rural settings where bandwidth is scarce, compute is limited, and clinical decision-making requires integrating heterogeneous modalities. We introduce a cloud--edge collaborative architecture that addresses these constraints: lightweight, domain-specific models on the edge transform raw medical data into compact structured outputs, while a cloud LLM synthesizes these outputs into clinical summaries. An LLM-based orchestrator dynamically selects diagnostic tools based on patient context, promoting relevant modality coverage without processing irrelevant inputs. We evaluate on 100 multimodal clinical cases spanning cardiac, obstetric, trauma, ophthalmology, and screening scenarios---including sparse-input presentations with missing modalities and dense-input presentations with many overlapping inputs---under three simulated network profiles (500\,kbps--5\,Mbps), reporting 95\% confidence intervals throughout. The hybrid system attains the highest oracle accuracy (0.87--0.90) and the strongest factual grounding (KG precision up to 0.96), together with high coverage precision (0.95--0.99), while transmitting only ${\sim}$6.5\,KB of structured evidence to the cloud---three orders of magnitude less than cloud-only baselines. It maintains bandwidth-invariant latency (25--38\,s) at up to 15$\times$ lower token cost. These results highlight the role of architectural design in improving evidence selectivity and factual grounding, rather than merely reducing upload size, under deployment constraints.
\end{abstract}
\begin{keywords}
  rural healthcare, edge AI, multi-agent systems, clinical decision support, telemedicine
\end{keywords}
\section{Introduction}
\begin{figure}[t]
\centering
\includegraphics[width=\linewidth]{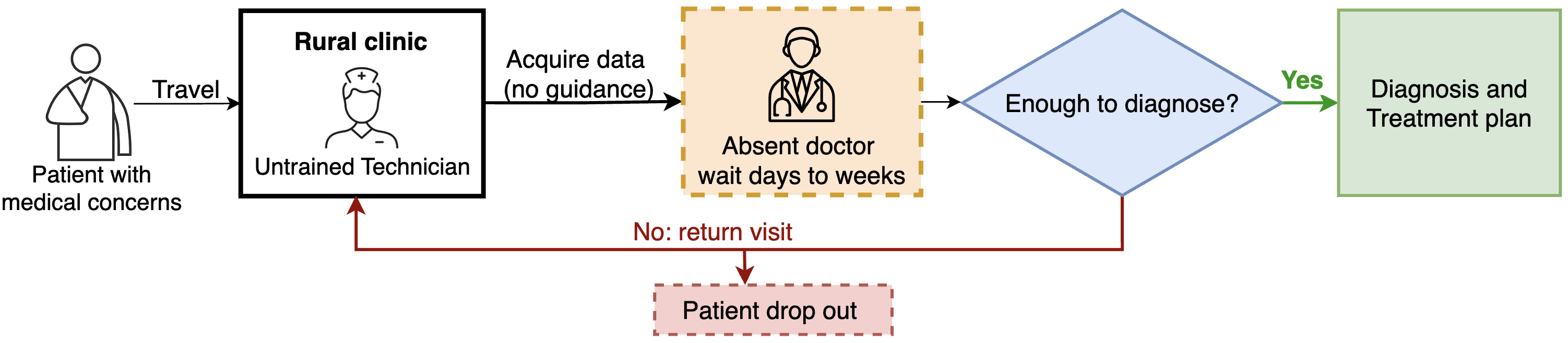}
\caption{Current rural diagnostic workflow. A patient travels to a rural clinic staffed by a technician without on-site specialist expertise, who acquires whichever modalities appear clinically relevant. The patient then waits days to weeks for remote specialist feedback. If modality coverage is sufficient, diagnosis and treatment follow; otherwise, the patient is asked to return for an additional visit. Each round introduces attrition.}
\label{fig:problem}
\end{figure}

Medical AI can now match specialists in reading chest X-rays~\cite{rajpurkar2017chexnet}, classifying arrhythmias~\cite{hannun2019cardiologist}, and grading retinal diseases~\cite{gulshan2016development}. However, these advances have yet to reach the rural communities that need them most~\cite{brown2026gaps, hrsa2022, topchik2020rural}. The disconnect lies not in what AI can do, but in where and how clinical diagnosis happens.
Clinical diagnosis is inherently multimodal and multi-turn. As shown in Fig.~\ref{fig:problem}, a typical diagnosis/clinical workflow in the rural clinic starts from patient complaints, iteratively order and interpret diverse tests, and synthesize findings across modalities. Supporting this process in rural clinics imposes three fundamental constraints that existing systems fall short of addressing. First, the AI system must decide which tests to acquire and gather sufficient evidence within a single encounter, as follow-up visits are often delayed or unavailable due to long travel distances, unreliable transportation, and limited clinic availability~\cite{douthit2015exposing, syed2013traveling}. Moreover, approximately half of specialist referrals from rural clinics are never completed~\cite{biggerstaff2017referrals}. Second, it should reliably ingest and interpret diverse test modalities in the absence of on-site specialist oversight. Finally, it must operate under severe bandwidth and compute constraints, as rural broadband speeds often range from 0.5--25\,Mbps with frequent dropouts~\cite{fcc2023}.
Existing approaches address these constraints only partially.  Domain-specific medical models (such as carotid ultrasound~\cite{jiang2025towards}, fetal ultrasound~\cite{maani2025fetalclip,van2018automated}, chest radiograph~\cite{cohen2022torchxrayvision,irvin2019chexpert} and blood pressure readings~\cite{carey2018prevention}) achieve specialist-level accuracy on individual modalities and deploy efficiently on edge hardware, but they operate on single modalities in isolation and cannot decide what to acquire or synthesize findings across tests. In contrast, Multimodal LLMs can reason across heterogeneous inputs and in principle handle the synthesis step, but their scale places them in the cloud, requiring raw medical data to be uploaded for inference which is often infeasible under rural bandwidth constraints. Additionally, their perceptual accuracy on specialized medical modalities still lags task-specific expert models. These limitations highlight a fundamental gap: no existing system simultaneously supports efficient deployment, multimodal integration, and reliable clinical reasoning under real-world resource constraints.
In this work, we address this gap by introducing a \emph{cloud--edge collaborative multi-agent system} that combines lightweight, domain-specific models on the edge with centralized reasoning in the cloud. Edge models transform raw medical data into compact structured outputs, eliminating the need to transmit bandwidth-intensive inputs. An orchestrator on the edge dynamically selects which diagnostic tools to invoke, ensuring that clinically relevant modalities are acquired during the patient encounter. A cloud-based reasoning agent integrates these outputs into coherent clinical summaries and recommendations. This design leverages the complementary strengths of specialized perception models and general-purpose reasoning models, while respecting real-world deployment constraints.
We evaluate on 100 diverse multimodal clinical cases spanning cardiac, obstetric, trauma, ophthalmology, and screening presentations under three simulated network profiles. The system is competitive-to-best on clinical accuracy, achieves the strongest factual grounding among all configurations, and maintains bandwidth-invariant latency at significantly lower token cost. Critically, we find that the architectural boundary between perception and reasoning shapes not just efficiency but the character of clinical errors: hybrid systems produce more selective, better-grounded evidence and fewer hallucinations than both cloud-based agentic systems and multimodal LLMs reasoning directly over raw data.
 We make the following contributions:
\begin{itemize}
\item A novel cloud–edge collaborative architecture for multimodal clinical screening that leverages an edge orchestrator for selective test modality acquisition, domain-specific edge models for specialist-level interpretation, and a cloud LLM for cross-modal reasoning.
\item A novel evaluation framework for bandwidth-constrained clinical AI, comprising 100 multimodal cases with real diagnostic data (including real-world ophthalmology teaching cases), sparse- and dense-input stress settings, three simulated rural network profiles, and a four-axis clinical quality evaluation including an oracle metric, tool-coverage metrics, KG-verification, and reasoning-quality decomposition, reported with 95\% confidence intervals.
\item A novel analysis of how architecture shapes diagnostic selectivity, factual grounding, and failure mode characterization.
\end{itemize}
\paragraph{Generalizable Insights} We show that the architectural boundary between perception and reasoning is not merely an efficiency choice but improves factual grounding: constraining a cloud LLM to reason over structured edge-tool outputs, rather than raw medical data, keeps clinical claims grounded in verifiable evidence and yields substantially higher factual \emph{precision} than cloud-only reasoning. Clinically, this decoupling delivers high-precision, well-grounded modality coverage (coverage precision 0.95--0.99; KG precision up to 0.96) with bandwidth-invariant latency (25--38\,s) under rural connectivity, suggesting that low-bandwidth, resource-constrained clinical settings may benefit from hybrid architectures that confine LLMs to reasoning and delegate perception to specialized models at the edge.
\section{Related Work}
\paragraph{Telemedicine, Cloud, Edge, and Multimodal Medical AI}
Telemedicine enables remote diagnosis by transmitting patient data to centralized experts~\cite{clarke2015investigation}, with adoption accelerating during COVID-19~\cite{hollander2020virtually}, and cloud-based medical AI now reaches high accuracy across dermatology, radiology, and ophthalmology~\cite{esteva2017dermatologist,rajpurkar2017chexnet,gulshan2016development}---but these approaches transmit bandwidth-intensive raw data and assume reliable connectivity. Edge deployment of lightweight, domain-specific models enables low-latency, cost-efficient inference~\cite{hannun2019cardiologist,howard2017mobilenets,rieke2020future,irvin2019chexpert}, yet is limited to single-modality inference without cross-modal integration or higher-level reasoning. Multimodal vision--language systems integrate imaging and text for report generation and VQA~\cite{zhang2024multimodal}, but are typically restricted to image--text pairs, assume centralized processing, and do not generalize to heterogeneous clinical modalities under bandwidth constraints.
\paragraph{LLM-Orchestrated and Agent-Based Systems}
LLMs have been explored for tool use and multi-step reasoning~\cite{schick2023toolformer}. In the medical domain, MedAgent-Pro~\cite{wang2025medagent} plans and invokes specialized visual tools with RAG-retrieved guidelines; MedCoAct~\cite{zheng2025medcoact} uses confidence-aware doctor--pharmacist collaboration; and AgentClinic~\cite{schmidgall2024agentclinic} benchmarks sequential decision-making under incomplete information. However, these \emph{systems} run in the cloud and, where multimodal, assume a pre-collected input bundle---MedAgent-Pro, for instance, drops plan steps whose inputs are absent rather than requesting them. Our orchestrator instead assembles the evidence set \emph{during} the encounter, each invocation being an edge-side modality acquisition that expands the state across rounds (\S\ref{sec:orch})---\emph{acquisition-and-analysis} orchestration over a growing bundle rather than analysis routing over a fixed one, the operative regime when no remote specialist pre-curates the input. \citet{liu2026benchmarking} further report only modest agent accuracy gains (0.5--8.9\%) at 10--100$\times$ token cost and 2--3$\times$ latency, the cost profile our edge-local design targets.
As summarized in \Cref{tab:comparison}, no existing system simultaneously supports edge execution, multimodal integration, bandwidth-aware orchestration, and reliable clinical reasoning under deployment constraints; our evaluation isolates how each choice (edge preprocessing, dynamic tool selection, structured communication) contributes to clinical quality.
\begin{table}[t]
\centering
\caption{Comparison with prior work across general system capabilities.
  \ding{51}\,=\,full support, $\triangle$\,=\,partial, \ding{55}\,=\,absent.}
\label{tab:comparison}
\small
\setlength{\tabcolsep}{4pt}
\begin{tabular}{l c c c c c c}
\toprule
\textbf{System}
  & \textbf{Multi.}
  & \textbf{Tool}
  & \textbf{Clin.}
  & \textbf{Deploy.}
  & \textbf{Edge}
  & \textbf{Low} \\[-0.3em]
  &\textbf{modal}
  &\textbf{Use}
  &\textbf{Reas.}
  &
  &
  &\textbf{BW} \\
\midrule
Telemedicine~\cite{clarke2015investigation}
  & \ding{51} & \ding{55} & \ding{55} & \ding{51} & \ding{55} & $\triangle$ \\
Cloud Med.\ AI~\cite{rajpurkar2017chexnet}
  & \ding{55} & \ding{55} & $\triangle$ & \ding{51} & \ding{55} & \ding{55} \\
Edge AI~\cite{hannun2019cardiologist,howard2017mobilenets}
  & \ding{55} & \ding{55} & \ding{55} & \ding{51} & \ding{51} & \ding{51} \\
LLM Tool Use~\cite{schick2023toolformer}
  & $\triangle$ & \ding{51} & \ding{51} & $\triangle$ & \ding{55} & \ding{55} \\
MedAgent-Pro~\cite{wang2025medagent}
  & \ding{51} & \ding{51} & \ding{51} & $\triangle$ & \ding{55} & \ding{55} \\
MedCoAct~\cite{zheng2025medcoact}
  & $\triangle$ & \ding{51} & \ding{51} & $\triangle$ & \ding{55} & \ding{55} \\
AgentClinic~\cite{schmidgall2024agentclinic}
  & \ding{51} & \ding{51} & \ding{51} & \ding{55} & \ding{55} & \ding{55} \\
Multimodal VLMs~\cite{zhang2024multimodal}
  & $\triangle$ & \ding{55} & \ding{51} & \ding{51} & \ding{55} & \ding{55} \\
\midrule
\textbf{Ours (Hybrid)}
  & \ding{51} & \ding{51} & \ding{51} & \ding{51} & \ding{51} & \ding{51} \\
\bottomrule
\end{tabular}
\end{table}
\section{System Architecture}
\label{sec:architecture}
\subsection{Overview}
\label{sec:arch-overview}
\begin{figure}[t]
\centering
\includegraphics[width=\linewidth]{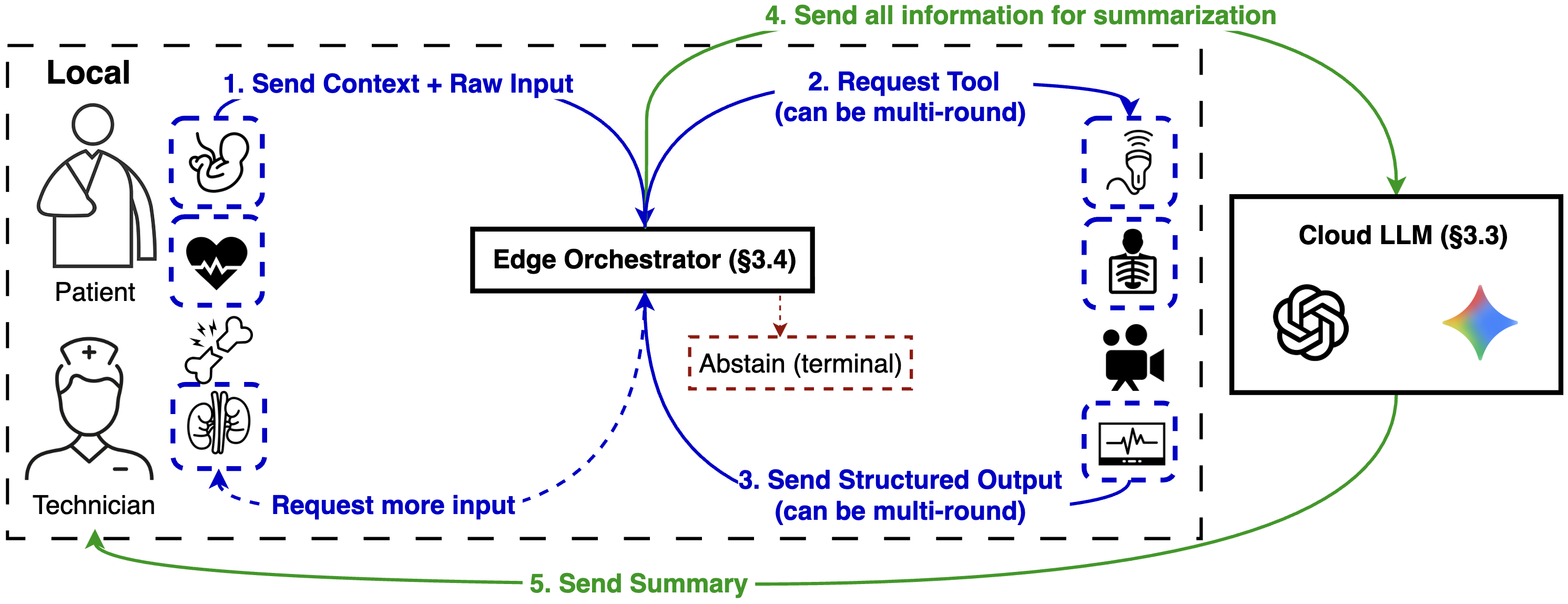}
\caption{Proposed cloud-edge system architecture. The local deployment (dashed outer box) couples a rural technician with the Edge Orchestrator~(\S3.4). During an encounter the orchestrator (1)~ingests raw acquisitions and patient context from the technician, (2)~requests additional tools from its repertoire when coverage is insufficient, optionally routing the request back to the technician for further on-patient acquisition (dashed), and (3)~receives structured outputs (findings, probabilities, key measurements) from each invoked tool, potentially over multiple rounds. If input quality or available evidence remains insufficient even after fallback, the orchestrator \emph{abstains} as a terminal action (red, dashed), closing the case with reacquisition guidance. Once evidence is sufficient, (4)~the orchestrator forwards the structured case record to the Cloud LLM~(\S3.3), which (5)~returns a diagnostic summary to the edge. Arrow color encodes communication locus: local tool-use (blue) vs.\ cloud round-trip (green). Raw acquisition and tool orchestration remain on-device.}
\label{fig:system-overview}
\end{figure}

The system comprises two layers connected by a lightweight communication protocol (\Cref{fig:system-overview}). The edge perception layer runs on local hardware and hosts specialized models for medical imaging and physiological signals, while the cloud reasoning layer synthesizes their outputs into clinical summaries using a large language model. An edge orchestrator selects which tools to invoke based on patient context and forwards results to the cloud once sufficient evidence has been gathered.
The key architectural invariant is that no raw image, video, or signal data crosses the edge--cloud boundary. All communication consists of structured JSON payloads (diagnostic labels, confidence scores, and quality flags), providing bandwidth guarantees independent of input size and restricting the cloud LLM to verifiable evidence.
\subsection{Edge Perception Layer}
\label{sec:edge}
The edge layer comprises seventeen specialized diagnostic tools spanning cardiac, obstetric, trauma/screening, and ophthalmic modalities: carotid ultrasound (UltraBot~\cite{jiang2025towards}), fetal ultrasound (FetalCLIP~\cite{maani2025fetalclip,van2018automated}), 12-lead ECG (ECGNet~\cite{hannun2019cardiologist,gow2023mimic}), chest radiograph (ChestXRay~\cite{cohen2022torchxrayvision,irvin2019chexpert}), thyroid ultrasound (ThyroidSeg~\cite{thyroid_segmentation_github}), CT scan (KidneyStone~\cite{yildirim2021deep}), extremity X-ray (BoneFracture~\cite{rajpurkar2017mura}), echocardiogram video (EchoNet~\cite{ouyang2020video}), facial video (BigSmall~\cite{narayanswamy2024bigsmall}), and blood pressure (BP~\cite{derdilla2026bp}), together with seven ophthalmology-oriented tools: retinal fundus grading, retinal OCT, fluorescein angiography vessel analysis, near-infrared retinal vessel imaging, slit-lamp assessment, B-scan ocular ultrasound, and a retinal vessel-segmentation grader. Each tool produces structured outputs with modality-specific fields, confidence scores, and quality flags.
Before forwarding outputs, the system applies deterministic quality gates. When segmentation quality is insufficient, MobileSAM~\cite{zhang2023faster} is invoked as a lightweight fallback. The registry is a modular, extensible interface rather than a closed clinical ontology: new modalities are incorporated through the same structured-output schema, though each added tool requires modality-specific validation, especially under rural acquisition conditions. Additional implementation details are provided in \Cref{app:implementation}.
\subsection{Cloud Reasoning Layer}
\label{sec:cloud}
The cloud layer receives accumulated evidence and patient context, including demographics, presenting symptoms, and medical history. The cloud LLM (Gemini or GPT; see \Cref{sec:baselines}) is prompted with these inputs together with structured tool outputs (diagnostic labels, confidence scores, and quality flags) and clinician notes.
The model synthesizes these inputs into a clinical summary that integrates findings across modalities, accounts for confidence signals, and produces diagnostic considerations and next-step recommendations. It operates exclusively on structured outputs and does not access raw data.
\subsection{Orchestrator}
\label{sec:orch}
The edge orchestrator is an LLM-based routing model that selects the next action given patient context, available tools, and accumulated evidence. It produces one of three decisions: \textbf{request-tool} (invoke additional tools, potentially in parallel), \textbf{summarize} (forward results to the cloud), or \textbf{abstain} (terminate with reacquisition guidance when evidence is insufficient).
\paragraph{Orchestrator Configuration.}
In each Hybrid configuration, the edge orchestrator is instantiated from
the same LLM family as the cloud synthesizer (Gemini~2.5~Pro or GPT-5.4),
invoked with JSON-mode decoding. Orchestrator calls are included in the token and latency totals reported in Table~\ref{tab:latency}. The orchestrator iteratively expands the evidence set and selects tools per case based on clinical relevance.
\section{Evaluation Framework}
We evaluate on 100 multimodal clinical cases spanning diverse patient demographics, acuity levels, and clinical presentations. Each case includes a patient profile (age, sex, symptoms, history), a set of relevant diagnostic inputs matched to the clinical context, and a set of red herring inputs from unrelated modalities designed to test tool selection specificity. The case set exercises the full range of edge tools and varies along several axes: number of relevant modalities, clinical urgency (routine screening to acute decompensation), and input density. Beyond the original cardiac/obstetric/trauma templates, the expanded benchmark adds \emph{sparse-input} presentations, in which clinically relevant modalities are missing, and \emph{dense-input} presentations, in which many relevant and irrelevant inputs are simultaneously available, to stress modality selection under both scarcity and clutter.
\paragraph{Case Generation}
Clinical scenarios are generated via few-shot LLM prompting with the tool registry included as context, inspired by the tool-aware prompt design of~\cite{ke2025dwim}. Case generation used Claude, whereas the evaluated systems use Gemini and GPT, so the generator does not overlap with the evaluated model families. Cases are populated with real, multi-source clinical data: ECG signals from MIMIC-IV-ECG Demo~\cite{gow2023mimic}, echocardiography videos from EchoNet-Dynamic~\cite{ouyang2020video}, blood pressure readings from MIMIC-IV Clinical Demo~\cite{johnson2023mimic}, and fetal ultrasound images from HC-18~\cite{van2018automated}. To move the evaluation beyond templated scenarios, the expanded benchmark also incorporates real-world cases, including multimodal ophthalmology case reports from EyeRounds.org, a public University of Iowa Department of Ophthalmology teaching resource. The ground truth for each case is defined as a set of verifiable findings used for automated oracle evaluation.
To mitigate potential bias from LLM-generated case designs, we note several safeguards. First, the diagnostic ground truth is derived from the real clinical data itself (dataset-annotated ejection fractions, clinically labeled ECG rhythms, known BP readings), not from the case generation LLM's assessment. Second, red-herring inputs are drawn from unrelated cases in the same source datasets. Third, the tool registry provided to the case generation LLM is the same registry available to all system configurations, so any case design bias would affect baselines equally. We nonetheless acknowledge that tool-registry conditioning can introduce tool-availability bias; the inclusion of real-world ophthalmology cases and sparse/dense settings is intended to reduce this dependence on the generation templates.
\subsection{Clinical Cases}
The 100 evaluation cases span ages 19--88 and cover cardiac and multi-system presentations (decompensated DCM, cardiogenic shock, aortic dissection, end-stage heart failure), obstetric presentations (severe preeclampsia, peripartum cardiomyopathy, HELLP), trauma/acute presentations (polytrauma, urosepsis), screening/controls (including a healthy negative control), and a 30-case ophthalmology bundle (acute macular neuroretinopathy, ocular toxoplasmosis, peripheral retinoschisis, acute retinal necrosis, and AZOOR/AIBSE). Cases test clinical complexity (ejection fractions 16.9--71.6\%, blood pressures from 72/44 to 208/120), tool selection specificity (red-herring inputs such as fetal ultrasound for a 68-year-old male), and cross-domain reasoning (e.g., connecting a urological source to cardiac dysfunction via sepsis). The full case table is provided in \Cref{app:cases}.
\subsection{Network Profiles}
\label{sec:network-profiles}
To evaluate system behavior under realistic connectivity constraints, we define three bandwidth profiles that model the uplink channel between the edge device and the cloud (\Cref{tab:network}). Bandwidth is modeled as a log-normal process with periodic updates, discrete dropout events, and hard floor/ceiling bounds, implemented as a Python-level socket throttle.
\begin{table}[t]
\centering
\caption{Network profiles for bandwidth-constrained evaluation. Bandwidth follows a log-normal random walk with the listed parameters, updated every $\Delta t$ seconds. Dropout events zero the link for the listed duration with probability $p$ per update.}
\label{tab:network}
\small
\begin{tabular}{lccccccc}
\toprule
Profile & Base & Floor & Ceiling & $\sigma$ & $\Delta t$ & $p_{\text{drop}}$ & Drop\,dur. \\
 & (kbps) & (kbps) & (kbps) & & (s) & & (s) \\
\midrule
Rural low & 500 & 20 & 2{,}000 & 0.8 & 3.0 & 0.15 & 5.0 \\
Rural moderate & 2{,}000 & 100 & 10{,}000 & 0.5 & 4.0 & 0.05 & 3.0 \\
Rural good & 5{,}000 & 500 & 20{,}000 & 0.3 & 5.0 & 0.02 & 2.0 \\
\bottomrule
\end{tabular}
\end{table}
\textbf{Rural low} models a degraded cellular or satellite uplink far from any tower, based on ITU measurements showing median rural upload speeds below 1\,Mbps in low-connectivity countries~\cite{itu2024facts}. \textbf{Rural moderate} represents US communities served by aging DSL, where the FCC found median upload speeds of 2--3\,Mbps~\cite{fcc2024fixedbroadband}. \textbf{Rural good} represents fixed wireless or early LEO satellite service, based on Ookla measurements of Starlink upload (14.8\,Mbps median) and rural fixed wireless~\cite{ookla2025starlink,ookla2023fwa}.
\subsection{Baselines}
\label{sec:baselines}
We evaluate six system configurations spanning two cloud LLM families (Gemini 2.5 Pro and GPT-5.4) and three architectural paradigms:
\begin{itemize}
    \item \textbf{Hybrid} (edge + cloud): The proposed system. Specialized models run locally on edge hardware, an LLM orchestrator selects which tools to invoke, and structured outputs are sent to the cloud LLM for synthesis. No raw images or signals cross the edge--cloud boundary.
    \item \textbf{Agentic}: Cloud-only multi-turn baseline. The cloud LLM receives patient context and a list of available inputs, then iteratively requests specific images and signals to examine across multiple rounds. Raw data is uploaded on demand.
    \item \textbf{Direct}: Cloud-only single-shot baseline. All available inputs (both relevant and red herrings) are uploaded to the cloud LLM simultaneously with patient context.
\end{itemize}
Each paradigm is tested with both LLMs for fair cross-model comparison. All six configurations use the same 100 clinical cases, network throttling profiles, and evaluation metrics.
\subsection{Accuracy Metrics}
\label{sec:accuracy-metrics}
We evaluate clinical quality along four complementary axes: (1)~\textbf{Oracle Accuracy}: micro-averaged accuracy of verifiable findings (BP category, ECG rhythm, EF range, CXR pathologies) extracted via pattern matching against ground truth; (2)~\textbf{Input Coverage}: recall (clinically relevant tools invoked) and precision (irrelevant tools avoided); (3)~\textbf{KG Verification}: medical claims verified against a MeSH-based knowledge graph through a multi-stage cascade (described below); (4)~\textbf{Reasoning Quality}: following MedR-Bench~\cite{qiu2025quantifying}, we decompose summaries into reasoning steps and evaluate efficiency, factuality, and completeness. We report 95\% confidence intervals for all metrics.
\paragraph{KG Verification Pipeline}
Claims are extracted from summary text using SciSpacy named entity recognition (the \texttt{en\_ner\_bc5cdr\_md} model trained on diseases and chemicals), then filtered through a stoplist, measurement regex, and negation detection. Each surviving mention is verified through a five-stage cascade, accepting the first stage that succeeds: (1)~structured verification against edge tool outputs; (2)~text overlap via synonym canonicalization and Jaccard overlap ($\geq$0.5); (3)~MeSH graph traversal using SapBERT embeddings~\cite{liu2021self} with FAISS retrieval (cosine $\geq$0.65) and $\leq$3 BFS hops; (4)~direct SapBERT embedding similarity ($\geq$0.55); (5)~RAG literature matching (cosine $\geq$0.50). Stages~1--2 handle \emph{echoed} claims; stages~3--5 handle \emph{novel} claims. We track both populations separately and classify unsupported claims as verification gaps, speculative, or genuine hallucinations. We present KG verification as a factual-grounding \emph{proxy}, not clinical adjudication; its similarity thresholds are heuristic rather than independently clinically validated.
\subsection{Efficiency Metrics}
We measure total end-to-end latency (decomposed into orchestration, tool execution, cloud LLM generation, and data upload), LLM token cost, and cloud data transmitted.
\section{Results}
\label{sec:results}
\Cref{tab:main} presents the primary comparison across all six system configurations and three network profiles (point estimates; full 95\% confidence intervals in \Cref{app:ci}). Each cell reports the mean across traces. The results reveal a consistent pattern: architecture determines not only efficiency but the character of clinical errors. We trace this pattern from input acquisition (\Cref{sec:results-coverage}) through factual grounding (\Cref{sec:results-factual}), per-modality accuracy (\Cref{sec:results-modality}), reasoning quality (\Cref{sec:results-reasoning}), and efficiency (\Cref{sec:results-latency}).
\begin{table}[t]
\centering
\caption{Main system comparison across network profiles (point estimates; 95\% CIs in \Cref{app:ci}). Best value per metric per profile in \textbf{bold}. Coverage recall for the Direct baselines is \emph{definitional}: they upload all available inputs, so their recall is trivially near-perfect; it is therefore excluded from bolding, and the bolded recall is the best among the \emph{selective} systems (Hybrid, Agentic). Cloud Bytes: KB for Hybrid, MB otherwise. Reasoning-quality metrics (efficiency, factuality, completeness) are reported in \S\ref{sec:results-reasoning}.}
\label{tab:main}
\small
\setlength{\tabcolsep}{4pt}
\begin{tabular}{llcccccc}
\toprule
& & \multicolumn{2}{c}{Hybrid (Ours)} & \multicolumn{2}{c}{Agentic} & \multicolumn{2}{c}{Direct} \\
\cmidrule(lr){3-4} \cmidrule(lr){5-6} \cmidrule(lr){7-8}
Category / Metric & & Gemini & GPT & Gemini & GPT & Gemini & GPT \\
\midrule
\multicolumn{8}{l}{\textbf{Rural Low} (500\,kbps base)} \\
\quad Oracle Accuracy & & 0.884 & \textbf{0.902} & 0.703 & 0.812 & 0.823 & 0.861 \\
\quad Coverage & Recall & 0.742 & \textbf{0.771} & 0.598 & 0.752 & 0.976 & 1.000 \\
& Precision & 0.948 & 0.969 & 0.973 & \textbf{0.986} & 0.566 & 0.572 \\
\quad KG Verif.\ & Precision & \textbf{0.949} & 0.905 & 0.742 & 0.836 & 0.684 & 0.812 \\
& F1 & \textbf{0.766} & 0.711 & 0.694 & 0.681 & 0.641 & 0.761 \\
\quad Efficiency & Latency (s) & 38.2 & \textbf{29.8} & 56.4 & 91.7 & 148.6 & 119.8 \\
& Token cost & 8{,}146 & \textbf{1{,}913} & 17{,}482 & 15{,}911 & 11{,}204 & 27{,}948 \\
& Cloud Bytes & 6.5\,K & \textbf{6.4\,K} & 2.15\,M & 2.15\,M & 6.86\,M & 7.33\,M \\
\midrule
\multicolumn{8}{l}{\textbf{Rural Moderate} (2\,Mbps base)} \\
\quad Oracle Accuracy & & \textbf{0.886} & 0.884 & 0.714 & 0.799 & 0.831 & 0.844 \\
\quad Coverage & Recall & \textbf{0.766} & 0.758 & 0.622 & 0.740 & 0.984 & 1.000 \\
& Precision & 0.961 & 0.977 & 0.989 & \textbf{1.000} & 0.570 & 0.571 \\
\quad KG Verif.\ & Precision & \textbf{0.957} & 0.921 & 0.771 & 0.857 & 0.712 & 0.836 \\
& F1 & 0.779 & 0.733 & 0.718 & 0.702 & 0.667 & \textbf{0.794} \\
\quad Efficiency & Latency (s) & 35.6 & \textbf{27.4} & 45.8 & 39.6 & 72.4 & 75.1 \\
& Token cost & 8{,}797 & \textbf{1{,}860} & 18{,}068 & 16{,}436 & 11{,}560 & 28{,}674 \\
& Cloud Bytes & 6.5\,K & \textbf{6.4\,K} & 2.21\,M & 2.21\,M & 6.93\,M & 7.36\,M \\
\midrule
\multicolumn{8}{l}{\textbf{Rural Good} (5\,Mbps base)} \\
\quad Oracle Accuracy & & 0.871 & \textbf{0.879} & 0.713 & 0.745 & 0.822 & 0.863 \\
\quad Coverage & Recall & \textbf{0.789} & 0.753 & 0.662 & 0.740 & 0.994 & 1.000 \\
& Precision & 0.956 & 0.977 & 0.976 & \textbf{1.000} & 0.569 & 0.571 \\
\quad KG Verif.\ & Precision & \textbf{0.954} & 0.935 & 0.770 & 0.854 & 0.848 & 0.839 \\
& F1 & 0.789 & 0.740 & 0.706 & 0.697 & \textbf{0.806} & 0.782 \\
\quad Efficiency & Latency (s) & 33.1 & \textbf{25.2} & 38.4 & 31.6 & 59.7 & 64.1 \\
& Token cost & 9{,}844 & \textbf{1{,}829} & 20{,}276 & 15{,}137 & 14{,}589 & 28{,}646 \\
& Cloud Bytes & 6.5\,K & \textbf{6.4\,K} & 2.30\,M & 2.30\,M & 7.32\,M & 7.36\,M \\
\bottomrule
\end{tabular}
\end{table}
\subsection{Tool Invocation Coverage}
\label{sec:results-coverage}
The first question is whether each architecture acquires the diagnostic data needed for a complete clinical assessment---and, equally, whether it avoids ingesting irrelevant data. \textbf{Direct} attains near-perfect coverage recall (0.98--1.00), but this is definitional rather than a merit: it uploads \emph{every} available input, so it trivially receives all relevant modalities---along with all red herrings, which collapses its coverage precision to ${\sim}$0.57 and inflates its cloud payload by three orders of magnitude (\Cref{sec:results-latency}). Recall is therefore not a meaningful axis of comparison for Direct; the informative comparison is among the \emph{selective} architectures.
Between the two selective systems, \textbf{Hybrid} and \textbf{Agentic} attain comparable, uniformly high coverage precision (Hybrid 0.95--0.98, Agentic 0.97--1.00), confirming that both successfully avoid irrelevant modalities. The decisive difference is recall: the \textbf{agentic baseline consistently under-acquires}, recovering only 0.60--0.75 of the relevant modalities, whereas the Hybrid orchestrator recovers 0.74--0.79. The gap is substantial in aggregate (mean recall 0.76 vs.\ 0.69) and largest for the Gemini family, where the agentic baseline misses roughly a third of the relevant modalities (recall 0.60--0.66; e.g., 0.662 vs.\ Hybrid's 0.789 under Rural Good); Hybrid also stays ahead for the GPT family, by a smaller margin (e.g., 0.771 vs.\ 0.752 under Rural Low). At matched precision, then, an LLM left to request inputs on its own tends to stop early and leave relevant modalities unacquired, while the edge orchestrator assembles a more complete evidence set. This coverage advantage compounds with the much larger Hybrid lead in factual grounding and oracle accuracy (\Cref{sec:results-factual}). Finally, Hybrid preserves all tool outputs locally on the edge device, so structured diagnostic evidence remains available even if the uplink fails.
\subsection{Factual Quality}
\label{sec:results-factual}
Given the data each system collects, the next question is whether its clinical claims are grounded in that evidence or extrapolated beyond it. Hybrid systems achieve the highest KG verification \emph{precision} across all network profiles. Gemini Hybrid leads (0.949--0.957, stable across profiles), followed by GPT Hybrid (0.905--0.935). Cloud-only baselines are substantially lower: Agentic 0.742--0.857 and Direct 0.684--0.848. This gap reflects a fundamental architectural advantage: Hybrid systems reason over structured tool outputs with explicit confidence scores, constraining the LLM to claims grounded in verifiable evidence.
The KG \emph{F1} picture is more nuanced and we report it honestly: because Direct baselines ingest every input, their high recall of echoed findings lifts their F1, so Direct occasionally matches or exceeds Hybrid on F1 (e.g., Direct-GPT 0.794 vs.\ Gemini Hybrid 0.779 under Rural Moderate; Direct-Gemini 0.806 vs.\ Gemini Hybrid 0.789 under Rural Good). The consistent Hybrid advantage is in precision---i.e., not fabricating claims---rather than in recall of every mentionable finding.
Oracle accuracy tells a complementary story. Hybrid attains the highest oracle accuracy overall (Gemini 0.871--0.886, GPT 0.879--0.902). Among cloud-only baselines, GPT Direct is closest (0.844--0.863) but at up to 15$\times$ the token cost and 2--5$\times$ the latency; agentic baselines trail further (0.703--0.812).
\paragraph{Error Analysis: What Gets Hallucinated.}
To understand the precision gap, we categorize unsupported claims by their source. From manual review across configurations, we observed four recurring patterns.
\emph{Drug and protocol recommendations.} Cloud-only systems frequently generate specific treatment recommendations, such as drug names, dosages, lab panels, that are clinically plausible but not grounded in the patient's diagnostic data. In the preeclampsia case (case~03), Gemini Agentic produces 14 unsupported claims including specific medications (labetalol, hydralazine, nicardipine, magnesium sulfate), laboratory tests (creatinine, uric acid, proteinuria), and prognostic warnings (stroke risk, seizure prophylaxis). The Hybrid system avoids this pattern because its summary prompt constrains the LLM to synthesize structured tool outputs rather than generate treatment plans from training data.
\emph{Visual over-interpretation.} Direct-mode systems generate specific radiological or signal findings from unstructured visual interpretation. In the aortic dissection case (case~13), GPT Direct reports ``atherosclerotic calcification'' on chest X-ray and ``sinus tachycardia'' from a single uncalibrated ECG channel. Hybrid systems are architecturally immune: the LLM never sees raw images.
\emph{Context leakage.} Agentic systems that request too few inputs compensate by treating patient context (demographics, symptoms, history) as diagnostic findings. In case~13, an agentic run that requested only 2 inputs wrote ``Marfan syndrome,'' ``bicuspid aortic valve,'' and ``unequal arm blood pressures'' into its diagnosis as if these were findings from diagnostic data, when they were merely echoed from the patient history.
\emph{Speculative clinical reasoning.} Cloud-only systems generate plausible differential diagnoses or complications not supported by available evidence (``end-organ dysfunction,'' ``acute cardiac complications''). These are both easier to identify and less likely to mislead than perceptual or pharmacological hallucinations.
\subsection{Per-Modality Diagnostic Accuracy}
\label{sec:results-modality}
To understand where the factual quality gap originates, we examine accuracy on individual diagnostic modalities where ground-truth labels exist (full results in \Cref{app:modality}). Both Hybrid variants achieve near-perfect accuracy on modalities with specialized edge tools: 100\% for BP extraction, carotid stenosis, and thyroid nodule detection; 82--91\% for bone fracture. Cloud-only systems show substantially lower accuracy, particularly for bone fracture detection (Gemini Agentic: 62\% accuracy at 60\% coverage) and carotid stenosis. ECG classification accuracy is consistent across all systems (83--86\%), suggesting signal difficulty dominates over architecture. These results confirm that edge tool pre-processing provides the most reliable diagnostic extraction for modalities where specialized models exist.
\subsection{Reasoning Quality}
\label{sec:results-reasoning}
Beyond individual findings, clinical utility requires coherent diagnostic reasoning. We evaluate reasoning efficiency, factuality, and completeness following MedR-Bench. \emph{Reasoning-quality metrics are being recomputed on the full 100-case benchmark and will be reported here (placeholders in \Cref{tab:main}).} On the earlier case set, GPT Agentic achieved the highest reasoning factuality at much higher token cost, while GPT Hybrid achieved the highest completeness at competitive efficiency; we will confirm whether these trends persist at scale.
\subsection{Latency and Bandwidth}
\label{sec:results-latency}
\begin{figure*}[t]
\centering
\includegraphics[width=\textwidth]{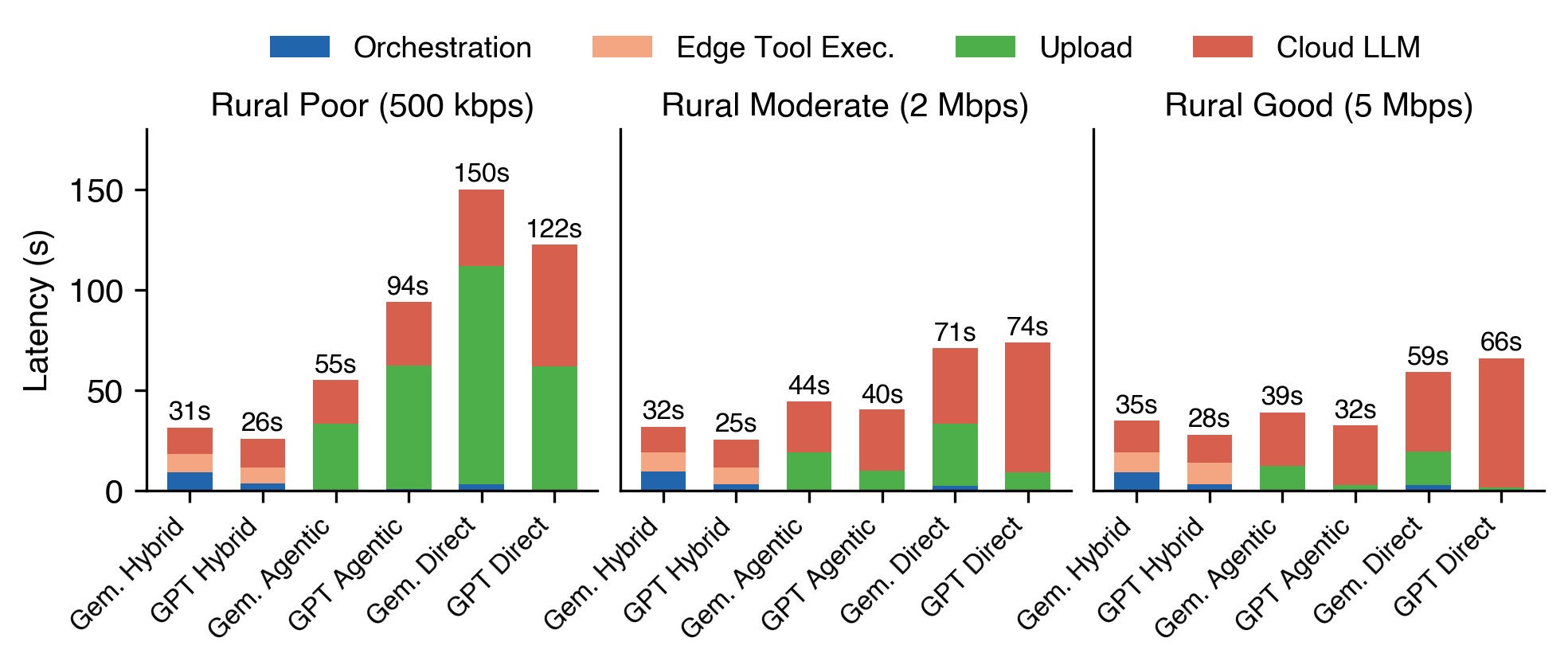}
\caption{Stacked latency breakdown under Rural Low (500\,kbps, left), Rural Moderate (2\,Mbps, center), and Rural Good (5\,Mbps, right)}
\label{fig:latency}
\end{figure*}

Finally, we evaluate whether quality differences come at a cost in deployment efficiency. \Cref{tab:latency} presents latency and token cost across configurations and network profiles.
\begin{table}[t]
\centering
\caption{Mean latency (seconds) and token cost across network profiles (point estimates; CIs in \Cref{app:ci}). Hybrid latency is bandwidth-invariant; cloud-only baselines show 1.4--3.9$\times$ slowdowns from good to poor connectivity.}
\label{tab:latency}
\small
\begin{tabular}{lcccccc}
\toprule
& \multicolumn{2}{c}{Rural Low} & \multicolumn{2}{c}{Rural Moderate} & \multicolumn{2}{c}{Rural Good} \\
\cmidrule(lr){2-3} \cmidrule(lr){4-5} \cmidrule(lr){6-7}
Config & Lat.\,(s) & Tokens & Lat.\,(s) & Tokens & Lat.\,(s) & Tokens \\
\midrule
Gemini Hybrid & 38.2 & 8{,}146 & 35.6 & 8{,}797 & 33.1 & 9{,}844 \\
GPT Hybrid & \textbf{29.8} & \textbf{1{,}913} & \textbf{27.4} & \textbf{1{,}860} & \textbf{25.2} & \textbf{1{,}829} \\
\midrule
Gemini Agentic & 56.4 & 17{,}482 & 45.8 & 18{,}068 & 38.4 & 20{,}276 \\
GPT Agentic & 91.7 & 15{,}911 & 39.6 & 16{,}436 & 31.6 & 15{,}137 \\
\midrule
Gemini Direct & 148.6 & 11{,}204 & 72.4 & 11{,}560 & 59.7 & 14{,}589 \\
GPT Direct & 119.8 & 27{,}948 & 75.1 & 28{,}674 & 64.1 & 28{,}646 \\
\bottomrule
\end{tabular}
\end{table}
Hybrid latency is essentially flat across bandwidth profiles (\Cref{fig:latency}): Gemini Hybrid varies only 33.1--38.2\,s and GPT Hybrid 25.2--29.8\,s. This stability arises because the Hybrid system transmits only ${\sim}$6.5\,KB of structured JSON, making bandwidth irrelevant. In contrast, cloud-only baselines transmit 2--7\,MB of raw data per case: Gemini Direct slows from 59.7\,s (rural good) to 148.6\,s (rural low), a 2.5$\times$ degradation. GPT Hybrid is the most token-efficient configuration at ${\sim}$1{,}860 tokens per case, a 15.4$\times$ reduction compared to GPT Direct (${\sim}$28{,}650 tokens). Gemini Hybrid's token savings are more modest (${\sim}$1.3--1.6$\times$ over Gemini Direct), so the largest token gains accrue to the GPT configuration; the bandwidth gain, by contrast, is uniform (${\sim}$1000$\times$) because it is set by the structured-output protocol rather than the LLM.
\section{Discussion}
\label{sec:discussion}
\paragraph{The Input Selection Problem.}
The agentic baselines reveal a second architectural insight: when an LLM controls its own data acquisition, it tends to under-acquire. Agentic systems recover only 60--75\% of clinically relevant modalities, and the consequences are not merely incomplete coverage but actively degraded reasoning. When a model lacks sufficient input evidence, it compensates by generating plausible claims from its training knowledge rather than acknowledging insufficient data. This compensation pattern is insidious because the resulting summaries read as confidently evidence-based.
The Hybrid orchestrator mitigates this failure mode by decoupling data acquisition from reasoning. The edge orchestrator operates over a structured tool registry with metadata about each tool's clinical applicability, and its decisions are informed by patient context without the distraction of simultaneously processing raw medical data. This separation raises relevant-modality recall to 74--79\% (vs.\ 60--75\% for agentic baselines) while keeping claims far better grounded (KG precision 0.905--0.957 vs.\ 0.742--0.857), a combination that directly translates to downstream summary quality. We note that Direct baselines attain a definitionally perfect recall by uploading everything, so their recall is not an informative point of comparison; it comes at the cost of precision (${\sim}$0.57 coverage precision, ${\sim}$0.57 KG precision for Gemini) and a three-orders-of-magnitude larger payload. The design goal is selective, grounded acquisition, not maximal ingestion.
\paragraph{Bandwidth Invariance as a Clinical Guarantee.}
From a deployment perspective, the most practically significant result is that Hybrid latency is effectively independent of bandwidth: 25--38 seconds across all three network profiles, compared to up to 2.5$\times$ degradation for cloud-only systems from rural good to rural low. This stability arises because the system transmits approximately 6.5\,KB of structured JSON regardless of input complexity, compared to 2--7\,MB of raw multimodal data for cloud-only systems. In rural clinical settings where connectivity is unpredictable, this invariance provides a reliability guarantee that cloud-only systems cannot match.
Furthermore, the Hybrid architecture offers a unique graceful degradation property: even if the cloud uplink fails entirely during a patient encounter, all edge tool outputs persist locally as structured data. A specialist reviewing the case later has access to the same diagnostic outputs that would have been synthesized into the cloud summary, enabling manual clinical reasoning without repeating the diagnostic workup.
\paragraph{Robustness Across LLM Families.} \label{par:robustness}
The Hybrid architecture outperforms its cloud-only counterpart within
both LLM families (Gemini~2.5~Pro and GPT-5.4) on factual grounding and
oracle accuracy, suggesting the gain is
not backbone-specific. The two Hybrid \emph{configurations} also differ
from each other: GPT Hybrid is faster and more token-efficient
(Table~\ref{tab:latency}), while Gemini Hybrid achieves higher
KG-verification precision (Table~\ref{tab:main}). Because the orchestrator
and synthesizer are instantiated from the same family in each
configuration, we report these differences at the configuration level
rather than attributing them to a specific pipeline stage
(see~\S\ref{sec:limitations}). With confidence intervals reported, several
cross-profile differences for a fixed model fall within overlapping
intervals, consistent with sampling noise rather than bandwidth-induced
reasoning-quality changes.
\subsection{Qualitative Case Study: Aortic Dissection with Expert Review}
\label{sec:case_study}
We illustrate the clinical impact of architectural differences through Case 13, a 55-year-old male with Marfan syndrome and bicuspid aortic valve presenting with tearing chest pain, unequal arm blood pressures, and diaphoresis—a textbook acute aortic dissection. Five modalities are relevant; blood pressure is 198/112 mmHg (hypertensive crisis). Both summaries were independently reviewed by a board-certified physician.
\definecolor{errorred}{RGB}{180,40,40}
\definecolor{correctgreen}{RGB}{20,120,55}
\definecolor{abstainblue}{RGB}{30,90,160}
\tcbset{
  casebox/.style={
    fonttitle=\bfseries\footnotesize,
    fontupper=\footnotesize,
    coltitle=white,
    boxrule=0.7pt,
    arc=2pt,
    left=5pt, right=5pt, top=4pt, bottom=4pt,
    boxsep=2pt,
    width=\linewidth,
  }
}
\begin{figure}[t]
\noindent
\begin{minipage}[t]{0.49\textwidth}
\begin{tcolorbox}[casebox,
  colback=red!3, colframe=errorred, colbacktitle=errorred,
  title={Undersampled run (2 of 5 modalities)},
  equal height group=G,
]
\textit{Key findings.}
Echocardiogram: dilated aortic root, intimal flap, aortic regurgitation, and \textcolor{errorred}{\sout{mild pericardial effusion}}. \ldots
Chest X-ray: widened mediastinum; \textcolor{errorred}{\sout{surgical clips in lower abdomen}}. \ldots
\textcolor{errorred}{[ECG not requested.] [BP not reported.] [Carotid not requested.]}
\textit{Diagnosis.} Acute Type~A dissection. \textcolor{errorred}{\sout{Pericardial effusion raises concern for rupture}}. \ldots
\textit{Next steps.} Target SBP 100--120, HR 60--70. CTA. Transfer. \ldots
\end{tcolorbox}
\end{minipage}%
\hfill
\begin{minipage}[t]{0.49\textwidth}
\begin{tcolorbox}[casebox,
  colback=green!3, colframe=correctgreen, colbacktitle=correctgreen,
  title={Full-invocation run (5 of 5 modalities)},
  equal height group=G,
]
\textit{Key findings.}
Echocardiogram: \textcolor{correctgreen}{\uline{intimal flap in ascending aorta}}, aortic regurgitation, LV dilation; \textcolor{correctgreen}{\uline{Type~A}}. \ldots
BP: \textcolor{correctgreen}{\uline{198/112\,mmHg, pulse 100\,bpm}}. \ldots
Chest X-ray: \textcolor{correctgreen}{\uline{widened mediastinum}}, cardiomegaly. \ldots
ECG: \textcolor{abstainblue}{\uline{ no diagnostic conclusion drawn}}.
Carotid: \textcolor{correctgreen}{\uline{assessed for malperfusion}}. \ldots
\textit{Diagnosis.} Acute Type~A dissection, justified by \textcolor{correctgreen}{\uline{symptoms, Marfan risk, intimal flap, widened mediastinum}}. \ldots
\textit{Next steps.} Target SBP 100--120, HR 60--70. \textcolor{correctgreen}{\uline{Arterial line}}, CTA, surgical consult. \ldots
\end{tcolorbox}
\end{minipage}
\caption{Condensed summaries for Case~13 (suspected aortic dissection, 55M), generated by the same cloud LLM under cloud agentic (left) and our architecture (right).
\textbf{Left:} Agentic run acquired only echocardiogram and chest X-ray.
\textcolor{errorred}{Red strikethrough} marks hallucinated findings or missed important test orders; \textcolor{errorred}{bracketed red} marks omitted modalities.
\textbf{Right:} Hybrid run acquired all five relevant modalities.
\textcolor{correctgreen}{Green underline} marks evidence-grounded claims absent or distorted in the weak summary;
\textcolor{abstainblue}{blue underline} marks an abstention.}
\label{fig:case13_diff}
\end{figure}
Figure~\ref{fig:case13_diff} juxtaposes condensed excerpts.
The Hybrid system invoked all five tools and produced a reasoning trajectory the reviewer characterized as "clear, accurate, and with organized facts." Three qualities stand out. First, every diagnostic claim traces to a specific tool output, with the dangerously elevated blood pressure prominently reported. Second, the system admits that it was not able to interpret the ECG data and draw conclusion, refusing to fabricate any rhythm classification. Third, findings are organized in a structured hierarchy that mirrors specialist reasoning.
The cloud-only agentic configuration requested only two of the five relevant modalities yet produced a summary of comparable length and confidence. The reviewer identified three critical failures. Most dangerous is a hallucinated "mild pericardial effusion", potentially altering management in ways that could harm the patient. The summary also omits the severely elevated blood pressure, the single most actionable finding for a rural technician initiating management before transfer. The reviewer described the output as analogous to "a less capable physician who is less well spoken, less organized, less reliable."
\section{Limitations}
\label{sec:limitations}
\paragraph{Simulated scenarios and case diversity.} Most of our 100 cases attach real diagnostic data (MIMIC-IV ECGs, EchoNet-Dynamic echocardiograms, HC-18 fetal ultrasounds) to synthetically constructed patient profiles, supplemented by real-world EyeRounds.org ophthalmology cases. While the diagnostic data is real, most scenarios and their ground-truth labels are not from actual encounters, and the case-generation LLM sees the tool registry and may favor tool-based reasoning; real cases and sparse/dense settings reduce but do not remove this bias. The 100 cases also do not span all rural presentations (dermatological and musculoskeletal complaints remain underrepresented), and the seventeen-tool registry is a modular interface, not a complete rural-care ontology. This is a feasibility benchmark under simulated constraints, not a claim of deployment readiness; validation on prospective rural data is necessary.
\paragraph{Edge-tool domain shift.} Per-modality accuracy (100\% for BP, carotid stenosis, and thyroid nodules; 82--91\% for bone fracture) reflects in-distribution performance, since the diagnostic data shares source corpora with the underlying models. It does not predict behavior under rural domain shift---different acquisition hardware, less-experienced operators, and underrepresented populations---so each tool, including the ophthalmology additions, needs independent modality-specific validation before deployment.
\paragraph{Metric and design caveats.} KG verification is a factual-grounding proxy, not clinical adjudication; its similarity thresholds (\S\ref{sec:accuracy-metrics}) are heuristic and unvalidated, so absolute values would shift with different thresholds, though relative ordering between configurations would likely hold. In every Hybrid configuration the orchestrator and synthesizer share an LLM family, so the family-level differences in \S\ref{par:robustness} cannot be attributed to a single stage; a factorial design crossing orchestrator and synthesizer families would be required. Finally, the Case~13 study (\S\ref{sec:case_study}) is a single-case, single-reviewer illustration of failure modes already quantified in \S\ref{sec:results-coverage}--\ref{sec:results-modality}, not an independent clinical evaluation; blinded multi-reviewer adjudication is left to future work.
\section{Conclusion}
\label{sec:conclusion}
We presented a cloud--edge architecture for multimodal clinical screening in resource-constrained rural settings. By combining edge-based specialized models with cloud-based reasoning over structured outputs, the system supports dynamic test selection and cross-modal synthesis while avoiding raw data transmission. Across 100 diverse clinical cases and three network conditions, the approach achieves the strongest factual grounding and competitive-to-best clinical accuracy with stable latency and substantially lower bandwidth and token cost than cloud-only baselines. These results highlight the importance of coordinating perception and reasoning under deployment constraints, and suggest that hybrid architectures are a practical path toward scalable multimodal clinical AI in low-resource settings.
\bibliography{references}
\newpage
\appendix
\section{Implementation Details}
\label{app:implementation}
\label{app:tools}
\paragraph{Edge Tool Specifications.}
The edge layer comprises seventeen specialized diagnostic tools, each wrapping a trained model behind a standardized interface that accepts modality-specific inputs and returns a unified structured output schema (\texttt{ToolStructuredOutput}). The original ten tools are:
\begin{enumerate}
    \item \textbf{UltraBot} (carotid ultrasound): Segments plaque regions and estimates stenosis severity ($<$50\%, 50--69\%, $\geq$70\%). Outputs include mask area (pixels), mask fraction, and stenosis confidence~\cite{jiang2025towards}.
    \item \textbf{FetalCLIP} (fetal ultrasound): Classifies fetal ultrasound planes into nine anatomical categories (abdomen, brain, femur, heart, kidney, lips/nose, profile, spine, cervix) with brain subplane classification and quality scoring~\cite{maani2025fetalclip,van2018automated}.
    \item \textbf{ECGNet} (12-lead ECG): Binary rhythm classification (normal vs.\ abnormal) with per-segment voting across the signal~\cite{hannun2019cardiologist}. Reports classification confidence and segment-level counts. Trained on MIMIC-IV-ECG~\cite{gow2023mimic}.
    \item \textbf{ChestXRay} (chest radiograph): Screens for 18 pathological findings (atelectasis, cardiomegaly, consolidation, edema, effusion, etc.) using a DenseNet backbone~\cite{cohen2022torchxrayvision,irvin2019chexpert}. Reports per-finding probabilities and flags findings exceeding a confidence threshold.
    \item \textbf{ThyroidSeg} (thyroid ultrasound): Segments thyroid nodules and reports detection status, mask statistics, and segmentation confidence~\cite{thyroid_segmentation_github}.
    \item \textbf{KidneyStone} (CT scan): Binary classification (kidney stone vs.\ normal) with classification confidence and per-class probabilities~\cite{yildirim2021deep}.
    \item \textbf{BoneFracture} (extremity X-ray): Classifies fracture status across bone types (elbow, hand, shoulder) with detection confidence~\cite{rajpurkar2017mura}.
    \item \textbf{EchoNet} (echocardiogram video): Estimates left ventricular ejection fraction (EF\%) and categorizes cardiac function (normal, mildly reduced, reduced, hyperdynamic)~\cite{ouyang2020video}.
    \item \textbf{BigSmall} (facial video): Extracts heart rate and respiratory rate from facial video using remote photoplethysmography (rPPG), with action unit detection for stress scoring~\cite{narayanswamy2024bigsmall}.
    \item \textbf{BP} (blood pressure readings): Categorizes blood pressure according to AHA guidelines~\cite{carey2018prevention} (normal, elevated, hypertension stage 1/2, hypertensive crisis) and assigns risk levels~\cite{derdilla2026bp}.
\end{enumerate}
To demonstrate extensibility of the registry through the same structured-output interface, we add seven ophthalmology-oriented edge tools:
\begin{enumerate}
    \setcounter{enumi}{10}
    \item \textbf{Fundus} (retinal fundus): Grades fundus photographs for retinal pathology and reports per-finding probabilities and image-quality flags.
    \item \textbf{OCT} (retinal OCT): Classifies macular/retinal findings from optical coherence tomography with confidence scoring.
    \item \textbf{FAVessel} (fluorescein angiography): Analyzes retinal vasculature and leakage patterns from fluorescein angiography.
    \item \textbf{NIRVessel} (near-infrared retinal): Extracts retinal vessel structure from near-infrared imaging.
    \item \textbf{SlitLamp} (slit-lamp): Assesses anterior-segment findings from slit-lamp imaging.
    \item \textbf{OcularUS} (B-scan ocular ultrasound): Reports posterior-segment findings from B-scan ocular ultrasound.
    \item \textbf{VesselSeg} (retinal vessel segmentation): Segments retinal vasculature as a grading aid, sharing the MobileSAM segmentation-fallback path.
\end{enumerate}
Each ophthalmology tool serves a distinct ocular modality (see the case bundle in \Cref{app:cases}); the additions require modality-specific validation under rural acquisition conditions.
\section{Edge Tool Execution Pipeline}
\label{edge_execution}
Once the orchestrator (§3.4) selects a set of tools to invoke, the edge runtime executes them subject to deterministic quality gates, a segmentation fallback, and memory-aware concurrent scheduling.
All tools share a unified output schema (\texttt{ToolStructuredOutput}) containing modality-specific fields, a generic quality score, and a list of clinical flags.
\paragraph{Confidence Gating and Quality Control.}
Before forwarding tool outputs to the cloud, the system applies deterministic quality gates. Segmentation tools (carotid, thyroid) are subject to minimum mask thresholds: carotid outputs are rejected if the segmentation mask contains fewer than 500 pixels or covers less than 1\% of the image area; thyroid outputs require at least 50 pixels and 0.1\% coverage. Classification tools include confidence scores that the cloud LLM can use to weight its reasoning.
\paragraph{MobileSAM Fallback.}
When a segmentation tool produces a low-quality mask that fails the quality gate, the system invokes MobileSAM~\cite{zhang2023faster}, a lightweight, ONNX-compatible segment-anything model, as a fallback. The fallback re-segments the region of interest and, if the result passes the quality gate, substitutes it for the original output. This mechanism provides graceful degradation without requiring cloud round-trips. Across the 100-case benchmark the specialist segmenters passed the quality gate, so this MobileSAM fallback was effectively never triggered (fewer than 1\% of carotid/thyroid segmentation calls).

\paragraph{Concurrent Execution.}
When the orchestrator requests multiple tools simultaneously, a memory-aware batch executor schedules them for concurrent execution. Tools are bin-packed into waves based on their estimated peak GPU memory (ranging from 5\,MB for BP categorization to 300\,MB for EchoNet and BigSmall), with a configurable memory budget per wave (default: 2\,GB). Tools within a wave execute in parallel via thread pools, exploiting PyTorch's GIL release during CUDA operations. All experiments were CPU-only and required no GPU, adding ${\sim}$4--5\,s per case; GPU execution is an optional latency boost.
\section{Full Confidence Intervals}
\label{app:ci}
\Cref{tab:ci-low,tab:ci-mod,tab:ci-good} report point estimates with 95\% confidence intervals for all metrics under each network profile.
\begin{table}[h]
\centering
\caption{Rural Low: point estimate [95\% CI].}
\label{tab:ci-low}
\scriptsize
\setlength{\tabcolsep}{3pt}
\resizebox{\textwidth}{!}{%
\begin{tabular}{lcccccc}
\toprule
Metric & Hybrid-Gem & Hybrid-GPT & Agentic-Gem & Agentic-GPT & Direct-Gem & Direct-GPT \\
\midrule
Oracle Acc & 0.884 [0.824, 0.944] & 0.902 [0.849, 0.955] & 0.703 [0.615, 0.791] & 0.812 [0.738, 0.886] & 0.823 [0.742, 0.904] & 0.861 [0.802, 0.920] \\
Cov.\ Recall & 0.742 [0.666, 0.818] & 0.771 [0.705, 0.837] & 0.598 [0.520, 0.676] & 0.752 [0.681, 0.823] & 0.976 [0.952, 1.000] & 1.000 [1.000, 1.000] \\
Cov.\ Precision & 0.948 [0.900, 0.996] & 0.969 [0.931, 1.000] & 0.973 [0.941, 1.000] & 0.986 [0.960, 1.000] & 0.566 [0.534, 0.598] & 0.572 [0.540, 0.604] \\
KG Precision & 0.949 [0.923, 0.975] & 0.905 [0.872, 0.938] & 0.742 [0.684, 0.800] & 0.836 [0.792, 0.880] & 0.684 [0.552, 0.816] & 0.812 [0.776, 0.848] \\
KG F1 & 0.766 [0.736, 0.796] & 0.711 [0.678, 0.744] & 0.694 [0.654, 0.734] & 0.681 [0.638, 0.724] & 0.641 [0.520, 0.762] & 0.761 [0.735, 0.787] \\
Latency (s) & 38.2 [35.1, 41.3] & 29.8 [27.2, 32.4] & 56.4 [50.8, 62.0] & 91.7 [82.6, 100.8] & 148.6 [136.2, 161.0] & 119.8 [109.4, 130.2] \\
Token Cost & 8{,}146 [7{,}214, 9{,}078] & 1{,}913 [1{,}690, 2{,}036] & 17{,}482 [14{,}896, 20{,}068] & 15{,}911 [13{,}600, 18{,}628] & 11{,}204 [8{,}931, 13{,}477] & 27{,}948 [25{,}000, 32{,}274] \\
Cloud Bytes & 6{,}506 [6{,}398, 6{,}614] & 6{,}421 [6{,}306, 6{,}536] & 2.15M [1.58M, 2.66M] & 2.15M [1.60M, 2.70M] & 6.86M [6.01M, 7.71M] & 7.33M [6.65M, 8.01M] \\
\bottomrule
\end{tabular}}
\end{table}
\begin{table}[h]
\centering
\caption{Rural Moderate: point estimate [95\% CI].}
\label{tab:ci-mod}
\scriptsize
\setlength{\tabcolsep}{3pt}
\resizebox{\textwidth}{!}{%
\begin{tabular}{lcccccc}
\toprule
Metric & Hybrid-Gem & Hybrid-GPT & Agentic-Gem & Agentic-GPT & Direct-Gem & Direct-GPT \\
\midrule
Oracle Acc & 0.886 [0.834, 0.938] & 0.884 [0.835, 0.934] & 0.714 [0.627, 0.801] & 0.799 [0.731, 0.866] & 0.831 [0.757, 0.905] & 0.844 [0.788, 0.899] \\
Cov.\ Recall & 0.766 [0.694, 0.837] & 0.758 [0.696, 0.821] & 0.622 [0.547, 0.697] & 0.740 [0.670, 0.810] & 0.984 [0.966, 1.000] & 1.000 [1.000, 1.000] \\
Cov.\ Precision & 0.961 [0.916, 1.000] & 0.977 [0.942, 1.000] & 0.989 [0.966, 1.000] & 1.000 [0.975, 1.000] & 0.570 [0.539, 0.601] & 0.571 [0.539, 0.603] \\
KG Precision & 0.957 [0.935, 0.979] & 0.921 [0.891, 0.950] & 0.771 [0.721, 0.822] & 0.857 [0.814, 0.900] & 0.712 [0.580, 0.844] & 0.836 [0.804, 0.867] \\
KG F1 & 0.779 [0.754, 0.805] & 0.733 [0.703, 0.762] & 0.718 [0.684, 0.752] & 0.702 [0.663, 0.741] & 0.667 [0.545, 0.790] & 0.794 [0.775, 0.813] \\
Latency (s) & 35.6 [32.8, 38.4] & 27.4 [24.9, 29.9] & 45.8 [41.0, 50.6] & 39.6 [35.7, 43.5] & 72.4 [66.1, 78.7] & 75.1 [68.4, 81.8] \\
Token Cost & 8{,}797 [7{,}899, 9{,}695] & 1{,}860 [1{,}690, 2{,}030] & 18{,}068 [15{,}502, 20{,}634] & 16{,}436 [14{,}000, 18{,}872] & 11{,}560 [9{,}287, 13{,}832] & 28{,}674 [25{,}000, 32{,}348] \\
Cloud Bytes & 6{,}505 [6{,}401, 6{,}610] & 6{,}416 [6{,}302, 6{,}529] & 2.21M [1.68M, 2.73M] & 2.21M [1.68M, 2.73M] & 6.93M [6.07M, 7.79M] & 7.36M [6.68M, 8.04M] \\
\bottomrule
\end{tabular}}
\end{table}
\begin{table}[h]
\centering
\caption{Rural Good: point estimate [95\% CI].}
\label{tab:ci-good}
\scriptsize
\setlength{\tabcolsep}{3pt}
\resizebox{\textwidth}{!}{%
\begin{tabular}{lcccccc}
\toprule
Metric & Hybrid-Gem & Hybrid-GPT & Agentic-Gem & Agentic-GPT & Direct-Gem & Direct-GPT \\
\midrule
Oracle Acc & 0.871 [0.806, 0.935] & 0.879 [0.835, 0.924] & 0.713 [0.637, 0.788] & 0.745 [0.658, 0.831] & 0.822 [0.710, 0.935] & 0.863 [0.813, 0.914] \\
Cov.\ Recall & 0.789 [0.739, 0.838] & 0.753 [0.691, 0.815] & 0.662 [0.592, 0.732] & 0.740 [0.670, 0.810] & 0.994 [0.984, 1.000] & 1.000 [1.000, 1.000] \\
Cov.\ Precision & 0.956 [0.911, 1.000] & 0.977 [0.942, 1.000] & 0.976 [0.942, 1.000] & 1.000 [0.975, 1.000] & 0.569 [0.537, 0.601] & 0.571 [0.539, 0.603] \\
KG Precision & 0.954 [0.931, 0.976] & 0.935 [0.913, 0.957] & 0.770 [0.702, 0.838] & 0.854 [0.824, 0.884] & 0.848 [0.770, 0.927] & 0.839 [0.790, 0.888] \\
KG F1 & 0.789 [0.769, 0.808] & 0.740 [0.718, 0.762] & 0.706 [0.652, 0.759] & 0.697 [0.667, 0.727] & 0.806 [0.741, 0.871] & 0.782 [0.749, 0.815] \\
Latency (s) & 33.1 [30.5, 35.7] & 25.2 [23.0, 27.4] & 38.4 [34.9, 41.9] & 31.6 [28.5, 34.7] & 59.7 [54.8, 64.6] & 64.1 [58.3, 69.9] \\
Token Cost & 9{,}844 [9{,}381, 10{,}308] & 1{,}829 [1{,}660, 1{,}998] & 20{,}276 [16{,}707, 23{,}845] & 15{,}137 [12{,}900, 17{,}374] & 14{,}589 [13{,}355, 15{,}823] & 28{,}646 [25{,}000, 32{,}292] \\
Cloud Bytes & 6{,}504 [6{,}399, 6{,}609] & 6{,}399 [6{,}290, 6{,}508] & 2.30M [1.80M, 2.79M] & 2.30M [1.80M, 2.79M] & 7.32M [6.64M, 7.99M] & 7.36M [6.68M, 8.04M] \\
\bottomrule
\end{tabular}}
\end{table}
\section{Clinical Cases}
\label{app:cases}
\Cref{tab:cases-appendix} summarizes the 100 evaluation cases. Cardiac and multi-system cases include EchoNet-Dynamic echocardiography video (EF-matched to the clinical scenario) and MIMIC-IV-ECG signals; ophthalmology cases (071--100) use ocular-modality bundles. Rel.\ = number of relevant inputs. BP format: systolic/diastolic\,(mmHg); ``---'' denotes not applicable/unrecorded.
\begin{longtable}{llp{4.6cm}lcc}
\caption{Clinical evaluation cases (100).}\label{tab:cases-appendix}\\
\toprule
\# & Age/Sex & Primary Presentation & BP & Rel. & EF\,(\%) \\
\midrule
\endfirsthead
\multicolumn{6}{l}{\emph{(continued)}}\\
\toprule
\# & Age/Sex & Primary Presentation & BP & Rel. & EF\,(\%) \\
\midrule
\endhead
\bottomrule
\endfoot
\multicolumn{6}{l}{\emph{Heavy --- unique oracle-labelled cardiac cases (001--020)}} \\
001 & 68M & Decompensated DCM + renal calculi & 88/52 & 5 & 23.8 \\
002 & 74M & CHF exacerbation + persistent AFib & 148/88 & 6 & 39.8 \\
003 & 29F & Severe preeclampsia, 32\,wk & 172/110 & 6 & 56.3 \\
004 & 62M & Cardiogenic shock post-STEMI & 72/44 & 5 & 23.2 \\
005 & 34F & Peripartum cardiomyopathy, 36\,wk & 128/82 & 6 & 57.1 \\
006 & 42M & Motorcycle polytrauma, cardiac concern & 105/68 & 6 & 49.7 \\
007 & 24M & Sports cardiac screening (FHx SCD) & 118/72 & 5 & 71.6 \\
008 & 58M & Acute MI + renal colic & 168/98 & 6 & 38.5 \\
009 & 66M & Aortic stenosis + thyroid nodule & 142/78 & 6 & 61.0 \\
010 & 78M & Geriatric cardiac screening & 138/82 & 6 & 61.0 \\
011 & 31F & Postpartum thyroiditis & 132/74 & 5 & 41.6 \\
012 & 35M & Healthy negative control & 118/76 & 5 & 65.4 \\
013 & 55M & Suspected aortic dissection (Marfan) & 198/112 & 5 & 48.4 \\
014 & 49M & Urosepsis $\to$ septic cardiomyopathy & 78/42 & 6 & 43.3 \\
015 & 28F & Twin pregnancy, HELLP concern, 30\,wk & 182/116 & 6 & 64.3 \\
016 & 71M & End-stage HF, ICD, CKD-4 & 82/50 & 5 & 16.9 \\
017 & 63M & TIA + carotid stenosis 60\% + AFib & 156/92 & 6 & 69.4 \\
018 & 82M & Elderly fall on warfarin & 98/62 & 5 & 45.3 \\
019 & 33F & Late pregnancy + MVP decompensation & 144/88 & 6 & 55.4 \\
020 & 70M & Rheumatic multivalvular disease & 108/72 & 6 & 39.9 \\
\midrule
\multicolumn{6}{l}{\emph{Cardiac stress archetypes (021--030)}} \\
021 & 69M & CHF/COPD overlap dyspnea & 118/72 & 3 & --- \\
022 & 31F & Late-pregnancy HTN + headache & 138/82 & 2 & --- \\
023 & 78M & Anticoagulated fall, fracture vs.\ cardiac & 148/88 & 3 & --- \\
024 & 64M & TIA + carotid disease + neck mass & 104/66 & 5 & --- \\
025 & 56M & Urosepsis + secondary cardiac strain & 72/44 & 5 & --- \\
026 & 23M & Athlete exertional syncope screening & 148/88 & 5 & --- \\
027 & 35F & Postpartum dyspnea: thyroiditis vs.\ CMP & 128/82 & 5 & --- \\
028 & 44M & Polytrauma + shock + cardiac contusion & 172/110 & 7 & --- \\
029 & 82M & Dense geriatric multi-system screening & 88/52 & 9 & --- \\
030 & 52M & Renal colic + concurrent chest-pain risk & 88/52 & 7 & --- \\
\midrule
\multicolumn{6}{l}{\emph{Additional distinct cardiac / multi-system cases (031--070)}} \\
031 & 46F & Paroxysmal SVT, palpitations & 124/78 & 2 & 61.0 \\
032 & 59M & Hypertensive urgency & 198/116 & 2 & --- \\
033 & 38M & Recurrent renal colic & 138/84 & 2 & --- \\
034 & 44F & Solitary thyroid nodule & 126/80 & 2 & --- \\
035 & 71M & Asymptomatic carotid bruit & 152/88 & 2 & --- \\
036 & 67F & Fall, forearm fracture & 142/80 & 2 & --- \\
037 & 26F & Gestational hypertension, 28\,wk & 148/94 & 2 & --- \\
038 & 55M & Community-acquired pneumonia & 128/78 & 2 & --- \\
039 & 64M & Stable exertional angina & 138/82 & 2 & 55.4 \\
040 & 73F & New AFib with RVR & 132/86 & 2 & 48.4 \\
041 & 66M & COPD exacerbation & 134/82 & 2 & --- \\
042 & 49F & Syncope workup & 118/74 & 2 & 65.4 \\
043 & 33F & Thyrotoxic palpitations & 138/76 & 2 & --- \\
044 & 68M & Amaurosis fugax / TIA & 156/90 & 2 & --- \\
045 & 69M & HFrEF + persistent AFib & 118/74 & 4 & 39.8 \\
046 & 60M & NSTEMI + renal colic & 168/96 & 4 & 38.5 \\
047 & 30F & Preeclampsia evaluation, 33\,wk & 172/108 & 4 & 56.3 \\
048 & 66M & TIA: carotid disease + AFib & 150/88 & 4 & 61.0 \\
049 & 58F & Urosepsis + cardiac strain & 86/50 & 5 & 43.3 \\
050 & 21M & Athlete pre-participation screen & 120/74 & 4 & 69.4 \\
051 & 34F & Postpartum cardiomyopathy & 128/82 & 4 & 41.6 \\
052 & 65M & Aortic stenosis + thyroid mass & 142/78 & 4 & 61.0 \\
053 & 41M & Blunt chest trauma & 106/68 & 4 & 48.4 \\
054 & 77M & Geriatric cardiovascular screen & 138/80 & 5 & 61.0 \\
055 & 62F & Hypertensive heart disease & 178/102 & 4 & 55.4 \\
056 & 59M & Diabetic silent ischemia & 146/86 & 4 & 48.4 \\
057 & 70F & Cardiogenic vs infective dyspnea & 128/80 & 4 & 38.5 \\
058 & 72M & Pre-op carotid + cardiac risk & 148/84 & 4 & 61.0 \\
059 & 71M & Decompensated HF, multi-organ & 92/56 & 6 & 23.2 \\
060 & 44M & Polytrauma with shock & 84/48 & 6 & 45.3 \\
061 & 29F & Eclampsia crisis & 186/118 & 5 & 56.3 \\
062 & 56M & Septic shock, renal source & 78/44 & 5 & 43.3 \\
063 & 69M & Comprehensive stroke workup & 162/94 & 5 & 55.4 \\
064 & 58M & Transplant eval, end-stage HF & 82/50 & 5 & 16.9 \\
065 & 67M & Multivalvular disease + thyroid & 110/72 & 5 & 39.9 \\
066 & 82M & Elderly fall, multi-injury & 100/62 & 6 & 45.3 \\
067 & 35F & Peripartum multisystem illness & 138/86 & 5 & 57.1 \\
068 & 64M & Cardiorenal syndrome & 124/78 & 5 & 41.6 \\
069 & 57M & Hypertensive emergency, multi-organ & 208/120 & 5 & 48.4 \\
070 & 80M & Dense geriatric multi-system screen & 142/84 & 7 & 39.9 \\
\midrule
\multicolumn{6}{l}{\emph{Ophthalmology cases --- ocular modality bundle (071--100)}} \\
071 & 24F & Acute macular neuroretinopathy (OU) & --- & 4 & --- \\
072 & 31F & Acute macular neuroretinopathy (OD) & --- & 4 & --- \\
073 & 38F & Acute macular neuroretinopathy (OS) & --- & 4 & --- \\
074 & 27M & Acute macular neuroretinopathy (OU) & --- & 4 & --- \\
075 & 44F & Acute macular neuroretinopathy (OD) & --- & 4 & --- \\
076 & 19F & Acute macular neuroretinopathy (OS) & --- & 4 & --- \\
077 & 42M & Acquired ocular toxoplasmosis (OD) & --- & 3 & --- \\
078 & 29F & Acquired ocular toxoplasmosis (OS) & --- & 3 & --- \\
079 & 55M & Acquired ocular toxoplasmosis (OU) & --- & 3 & --- \\
080 & 36F & Acquired ocular toxoplasmosis (OD) & --- & 3 & --- \\
081 & 48M & Acquired ocular toxoplasmosis (OS) & --- & 3 & --- \\
082 & 23F & Acquired ocular toxoplasmosis (OD) & --- & 3 & --- \\
083 & 83M & Acquired peripheral retinoschisis (OU) & --- & 2 & --- \\
084 & 71F & Acquired peripheral retinoschisis (OD) & --- & 2 & --- \\
085 & 66M & Acquired peripheral retinoschisis (OS) & --- & 2 & --- \\
086 & 78F & Acquired peripheral retinoschisis (OU) & --- & 2 & --- \\
087 & 60M & Acquired peripheral retinoschisis (OD) & --- & 2 & --- \\
088 & 88F & Acquired peripheral retinoschisis (OS) & --- & 2 & --- \\
089 & 84M & Acute retinal necrosis (OD) & --- & 3 & --- \\
090 & 57M & Acute retinal necrosis (OS) & --- & 3 & --- \\
091 & 69F & Acute retinal necrosis (OU) & --- & 3 & --- \\
092 & 45M & Acute retinal necrosis (OD) & --- & 3 & --- \\
093 & 73F & Acute retinal necrosis (OS) & --- & 3 & --- \\
094 & 38M & Acute retinal necrosis (OD) & --- & 3 & --- \\
095 & 34F & AZOOR / AIBSE (OD) & --- & 3 & --- \\
096 & 28F & AZOOR / AIBSE (OS) & --- & 3 & --- \\
097 & 41F & AZOOR / AIBSE (OU) & --- & 3 & --- \\
098 & 22M & AZOOR / AIBSE (OD) & --- & 3 & --- \\
099 & 47F & AZOOR / AIBSE (OS) & --- & 3 & --- \\
100 & 31F & AZOOR / AIBSE (OD) & --- & 3 & --- \\
\end{longtable}
\section{Per-Modality Diagnostic Accuracy}
\label{app:modality}
\Cref{tab:modality-appendix} presents per-modality diagnostic accuracy across the six configurations, averaged over all three network profiles, on the subset of cases with verifiable ground-truth labels. (These per-modality values are being refreshed on the full 100-case set; the ophthalmology cases contribute to coverage but not to the label-verified modalities below.)
\begin{table}[h]
\centering
\caption{Per-modality diagnostic accuracy averaged across network profiles. Echo EF reports MAE and category accuracy. Other modalities report accuracy and coverage. Coverage here measures whether the modality's findings appear in the generated clinical summary, distinct from tool invocation coverage. Best per metric in \textbf{bold}.}
\label{tab:modality-appendix}
\small
\begin{tabular}{lcccccc}
\toprule
& \multicolumn{2}{c}{Hybrid (Ours)} & \multicolumn{2}{c}{Agentic} & \multicolumn{2}{c}{Direct} \\
\cmidrule(lr){2-3} \cmidrule(lr){4-5} \cmidrule(lr){6-7}
Metric & Gemini & GPT & Gemini & GPT & Gemini & GPT \\
\midrule
\multicolumn{7}{l}{\textbf{Echo EF}} \\
\quad MAE (\%) $\downarrow$ & 12.2 & 11.5 & 11.1 & \textbf{9.1} & 13.3 & 10.3 \\
\quad Category Acc.\ (\%) & 67.8 & 69.5 & \textbf{70.1} & 65.4 & 51.1 & 58.5 \\
\quad Coverage (\%) & \textbf{100.0} & \textbf{100.0} & 77.7 & 76.7 & 98.8 & 98.3 \\
\midrule
\multicolumn{7}{l}{\textbf{ECG Classification}} \\
\quad Accuracy (\%) & 84.3 & \textbf{85.7} & 82.8 & 85.1 & 83.6 & 84.7 \\
\quad Coverage (\%) & 95.8 & \textbf{99.2} & 79.1 & 95.0 & 79.2 & 98.3 \\
\midrule
\multicolumn{7}{l}{\textbf{Blood Pressure}} \\
\quad Accuracy (\%) & \textbf{100.0} & \textbf{100.0} & \textbf{100.0} & \textbf{100.0} & 93.8 & \textbf{100.0} \\
\quad Coverage (\%) & 92.5 & \textbf{100.0} & 67.3 & 86.7 & 72.7 & 98.3 \\
\midrule
\multicolumn{7}{l}{\textbf{Chest X-Ray Findings}} \\
\quad Accuracy (\%) & 77.8 & 77.9 & \textbf{82.2} & 80.4 & 75.6 & 74.6 \\
\quad Coverage (\%) & 90.0 & 94.2 & 72.1 & 85.0 & \textbf{98.9} & 98.3 \\
\midrule
\multicolumn{7}{l}{\textbf{Carotid Stenosis}} \\
\quad Accuracy (\%) & \textbf{100.0} & \textbf{100.0} & 93.9 & 88.0 & 81.2 & 78.4 \\
\quad Coverage (\%) & 96.2 & 82.7 & 47.1 & 48.1 & \textbf{100.0} & 98.1 \\
\midrule
\multicolumn{7}{l}{\textbf{Thyroid Nodule}} \\
\quad Accuracy (\%) & \textbf{100.0} & \textbf{100.0} & 63.6 & 90.9 & 62.1 & 91.7 \\
\quad Coverage (\%) & \textbf{100.0} & 95.8 & 35.5 & 45.8 & 96.7 & \textbf{100.0} \\
\midrule
\multicolumn{7}{l}{\textbf{Bone Fracture}} \\
\quad Accuracy (\%) & \textbf{90.9} & 81.8 & 61.5 & 71.4 & 56.7 & 47.4 \\
\quad Coverage (\%) & \textbf{100.0} & \textbf{100.0} & 59.6 & 41.8 & 64.3 & 86.4 \\
\multicolumn{7}{l}{\textbf{Color Fundus Photography}} \\
\quad Accuracy (\%) & 91.7 & \textbf{93.3} & 86.7 & 90.0 & 83.3 & 86.7 \\
\quad Coverage (\%) & \textbf{100.0} & \textbf{100.0} & 83.3 & 93.3 & 96.7 & \textbf{100.0} \\
\midrule
\multicolumn{7}{l}{\textbf{Optical Coherence Tomography (OCT)}} \\
\quad Accuracy (\%) & 93.3 & \textbf{95.0} & 88.3 & 91.7 & 85.0 & 88.3 \\
\quad Coverage (\%) & \textbf{100.0} & \textbf{100.0} & 86.7 & 95.0 & 98.3 & \textbf{100.0} \\
\midrule
\multicolumn{7}{l}{\textbf{Fundus Autofluorescence (FAF)}} \\
\quad Accuracy (\%) & 90.0 & \textbf{91.7} & 83.3 & 88.3 & 80.0 & 85.0 \\
\quad Coverage (\%) & 96.7 & \textbf{100.0} & 76.7 & 90.0 & 95.0 & 98.3 \\
\midrule
\multicolumn{7}{l}{\textbf{Fluorescein Angiography (FA)}} \\
\quad Accuracy (\%) & 88.9 & \textbf{91.7} & 83.3 & 87.5 & 79.2 & 83.3 \\
\quad Coverage (\%) & 95.8 & \textbf{100.0} & 75.0 & 87.5 & 95.8 & \textbf{100.0} \\
\bottomrule
\end{tabular}
\end{table}

\section{Privacy and Fully-Local Synthesis}
\label{app:privacy}
\paragraph{Privacy Considerations.}
Confining raw images and signals to the edge reduces exposure: no bandwidth-intensive pixel or waveform data leaves the clinic, and only compact structured outputs cross the edge--cloud boundary. This is a privacy \emph{benefit}, but not privacy by itself. The structured payloads still carry sensitive information---diagnostic labels, confidence scores, quality flags, and patient context---that can identify a patient or reveal a condition. A deployment therefore still requires standard protections on the uplink and the cloud service: transport and at-rest encryption of the structured records, authentication and role-based access control for the synthesis endpoint, and audit logging of every case submitted and every summary returned. We treat the edge--cloud protocol as reducing the attack surface (raw data never transits or persists in the cloud), not as a substitute for these controls; a full privacy and threat-model analysis, including re-identification risk from structured outputs, is left to future work.
\paragraph{Toward Fully Local Synthesis.}
Our design keeps only perception on the edge and delegates cross-modal reasoning to a cloud LLM, so the synthesis step still depends on connectivity (albeit at only ${\sim}$6.5\,KB per case). A natural extension is to run synthesis with a small, local language model (sLLM) on the same edge hardware, eliminating the cloud round-trip entirely and yielding summaries even under total uplink failure. Whether a compact on-device model can match the cross-modal reasoning quality of a cloud LLM over structured evidence---and at what latency on CPU-only rural hardware---is an open question we leave to future work.

\section{Reproducibility}
\label{app:repro}
We summarize the settings needed to reproduce our results and the artifacts we will release.
\paragraph{Models and decoding.} Cloud synthesizer and edge orchestrator: Gemini~2.5~Pro and GPT-5.4, invoked with JSON-mode decoding at temperature~0. The cloud summary uses a token budget of 256 (cap~1200); \texttt{top\_p} is left at the provider default.
\paragraph{Repetitions and confidence intervals.} Each case is run once per configuration and network profile (one trial). The 95\% CIs in \Cref{app:ci} are normal-approximate, $\bar{x}\pm 1.96\,\mathrm{SD}/\sqrt{n}$, with $n$ the number of evaluated cases (traces) per cell (metrics defined only on labelled findings use the corresponding subset).
\paragraph{Determinism and seeds.} The 100-case set is fixed. The bandwidth process is a stochastic log-normal random walk that is \emph{not} seeded across runs, so per-run latency/upload timings vary within the reported intervals; the CIs already reflect this together with case-level variance.
\paragraph{Case generation.} Cases were generated with Claude (Claude Code CLI) from a fixed DWIM-style clinical template with in-context examples (\Cref{app:prompts}); diagnostic ground truth comes from the source datasets (MIMIC-IV-ECG, EchoNet-Dynamic, MIMIC-IV, HC-18) and the EyeRounds.org reports, not the generator. Reasoning-quality metrics are scored by a separate Claude judge (\texttt{claude-haiku-4-5}).
\paragraph{Edge tool thresholds.} Segmentation gates: carotid mask $\geq$500\,px and $\geq$1\% area; thyroid $\geq$50\,px and $\geq$0.1\% area (\Cref{edge_execution}). Classifier report/flag thresholds: chest X-ray 0.5, 12-lead ECG 0.5, OCT 0.4, fundus 0.4 (diabetic-retinopathy gate 0.85), slit-lamp 0.4, B-scan ocular ultrasound 0.4; remaining tools emit calibrated heuristic confidences rather than a single hard threshold.
\paragraph{Network simulation.} Log-normal random-walk parameters are in \Cref{tab:network}; the socket-level throttle and dropout process will be released.
\paragraph{Evaluation.} Oracle pattern-matching, the five-stage KG cascade (\S\ref{sec:accuracy-metrics}), and the MedR-Bench reasoning decomposition are applied identically to all six configurations.
\paragraph{Artifact release.} We will release the generated case metadata, structured tool outputs, network-simulation scripts, and evaluation code.

\section{Prompts}
\label{app:prompts}
Runtime placeholders in braces (e.g.\ \{tools\_list\}, \{patient\_block\}, \{tool\_block\}) are substituted per case.

\subsection{Edge orchestrator (tool selection).}

\begin{lstlisting}
You are the EDGE ORCHESTRATOR for a rural clinic.
You DO NOT diagnose. You DO NOT write the final clinical summary.
Decide ONLY the next action:
- request_tool: run one of the available tools to gather evidence.
- summarize: evidence is sufficient; the cloud LLM writes the final summary.
- reacquire: input quality/evidence is insufficient; ask for more frames/images.
Available tools: {tools_list}
Heuristics:
- From age, symptoms, history, notes, decide which tool(s) are most valuable next.
- If several independent tools are clearly needed, request them ALL at once.
- Use "structured" to see gathered evidence; do NOT re-request a present tool.
- Once tools have produced results, almost always summarize (rural clinic;
  tests are hard to repeat); let the summarizer flag any inconsistency.
- "runnable_tools" lists tools whose inputs are present now; in the FIRST
  decision request ALL runnable tools unless clearly irrelevant.
Respond ONLY with a JSON object. Fields: decision
("request_tool"|"summarize"|"reacquire"), tools (list|null),
tool_args (object), reacquire_tips (list), reason (string).
\end{lstlisting}

\subsection{Cloud synthesis}

\begin{lstlisting}
You are a clinical decision-support assistant at a rural clinic. Given the
patient context and diagnostic tool results, write a clinical summary in prose.
RULES:
- REASON FROM HISTORY FIRST: distinctive clues (prior treatments, risk factors,
  symptom constellations) before tool results; tools CONFIRM/COMPLEMENT, not override.
- [HEURISTIC-ONLY - UNRELIABLE] = non-diagnostic noise.
- [LOW] reported but never primary; [MEDIUM] suggestive; [HIGH] strong but must
  be clinically consistent.
- Ground every claim in tool results or context; do NOT fabricate measurements.
- Do NOT mention modalities whose results are absent; "inconclusive" cannot
  support a diagnosis.
- The fundus tool is a diabetic-retinopathy grader ONLY; if the picture is
  non-diabetic, treat a DR label as a false positive.
- Label differentials "Possible"/"Consider"; never definitive. Prose only, no code.
{patient_block}
{tool_block}
Sections: KEY FINDINGS / DIFFERENTIAL DIAGNOSES / RECOMMENDED NEXT STEPS.
Now write the summary for the patient above.
\end{lstlisting}

\paragraph{Case generation.} Cases were generated from a fixed DWIM-style template (\texttt{prompt-clinical.xml}) with in-context clinical examples exposing a \texttt{ClinicalCase} schema (age/sex, symptoms, history, available input keys); the model emits a new multi-tool case as structured data, validated against the tool registry.

\paragraph{Agentic and Direct baselines.} Both baselines use the \emph{same} cloud-synthesis prompt above and differ only in data delivery: \textbf{Direct} uploads all available raw images/signals at once with the patient context, while \textbf{Agentic} uploads raw inputs iteratively, requesting specific modalities across rounds (Gemini/OpenAI file-upload APIs).
\end{document}